\documentclass[runningheads]{llncs}
\usepackage{graphicx}
\usepackage{amsmath,amssymb} 
\usepackage{color}
\usepackage{color}
\usepackage{cite}
\usepackage{subcaption}
\usepackage{xcolor}
\newcommand{\etal}{\textit{et al.}}
\begin{document}
\pagestyle{headings}
\mainmatter

\title{Robust Instance Tracking via Uncertainty Flow} 
\titlerunning{Robust Instance Tracking via Uncertainty Flow}
%
\author{Jianing Qian\inst{1} \and
Junyu Nan\inst{1}  \and Siddharth Ancha\inst{1} \and Brian Okorn\inst{1} \and David Held\inst{1}}
\authorrunning{Jianing, Junyu et al.}
%
\institute{Carnegie Mellon University\\
\email{\{jianingq,jnan1,sancha,bokorn,dheld\}@andrew.cmu.edu}}

\maketitle

\begin{abstract}
Current state-of-the-art trackers often fail due to distractors and large object appearance changes. In this work, we explore the use of dense optical flow to improve tracking robustness.  Our main insight is that, because flow estimation can also have errors, we need to incorporate an estimate of flow uncertainty for robust tracking.  We present a novel tracking framework which combines appearance and flow uncertainty information to track objects in challenging scenarios.  We experimentally verify that our framework improves tracking robustness, leading to new state-of-the-art results.  Further,  our experimental ablations shows the importance of flow uncertainty for robust tracking.
\end{abstract}

\section{Introduction}

Instance tracking is an important task in video applications, such as autonomous driving, sports analytics, video editing, and video surveillance. In single-object tracking, the position of the target instance is given in the first frame of a video sequence; tracking algorithms need to predict the position of the same instance in each of the following frames. 

Most state-of-the-art tracking methods use a convolutional neural network to extract features from the target object and features from the scene~\cite{SiamRPN,DaSiam,SiamMask}. 
These methods use an approach of ``tracking-by-one-shot-detection"~\cite{SiamRPN}: a network is trained to match the appearance between an image of the target object and an image of the same object in the current frame.

\begin{figure*}
\begin{center}
\includegraphics[width=\linewidth]{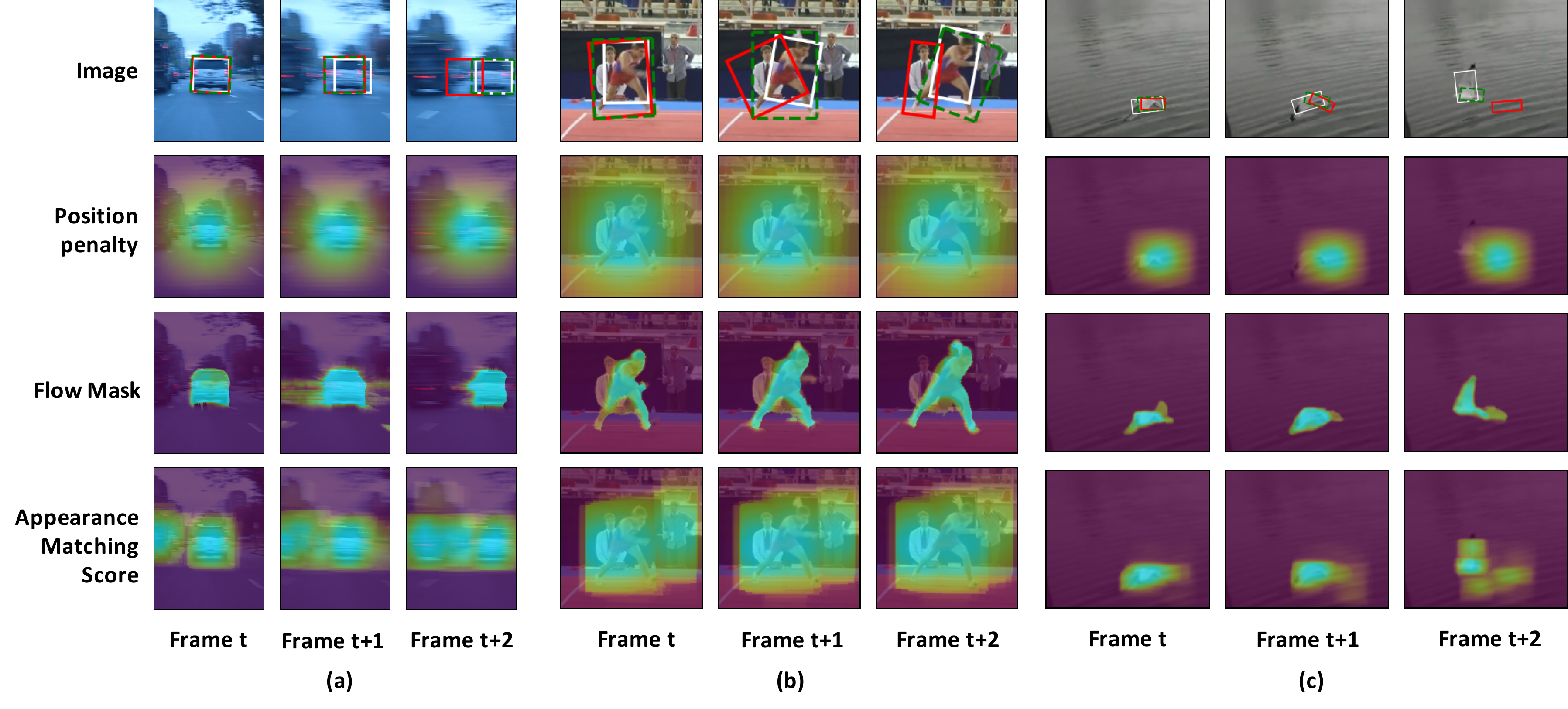}
\end{center}
   \caption{Three example errors that our method fixes (a): Failure case of baseline due to large camera motion; (b): Failure case of baseline due to distractors; (c): Failure case of baseline due to large motion of instance being tracked. In this figure, white boxes represent ground-truth boxes, \textcolor{red}{red} boxes represent predictions by SiamMask~\cite{SiamMask}, \textcolor{green}{green} boxes show the results from our method.}
\label{fig:all}
\end{figure*}

Although this ``tracking-by-one-shot-detection" approach has achieved impressive performance, it is prone to errors due to distractors and object appearance changes.  First, if there are similar-looking objects in the video (``distractors"), tracking-by-one-shot-detection methods often switch to a distractor object (see Figure~\ref{fig:all}b); common examples include different objects of the same category or objects of similar color or texture.
Likewise, if the object changes its appearance due to object deformations, image blur (from large camera or object motion), lighting changes, or other variations, tracking-by-one-shot-detection methods often lose track of the target object (see Figures~\ref{fig:all}a,~\ref{fig:all}c).  

When there are distractor objects or large appearance changes, matching the object appearance alone will likely be insufficient for robust tracking.  Instead, the tracker should make use of the tracked object's position.  By tracking the position of the object throughout the video, the tracker can determine which object is the target and which are the distractors. 

Past trackers have typically incorporated relatively weak position information to try to resolve these issues.  For example, many methods~\cite{SiamMask,SiamRPN,DaSiam, Fully_Convolutional_Siamese} use a position penalty that gives a lower score to bounding boxes that are farther away from the location of the detected object in the previous frame.  However, a position penalty that is too strong will lose track of fast moving objects or objects under large camera motion; a position penalty that is too weak will not achieve the desired effect of ignoring distractors or handling large object appearance changes.

To address these issues, we explore how trackers can incorporate object position information in a more robust manner.  Specifically, we use dense optical flow correspondences to track the position of the target object from one frame to the next.  Dense optical flow methods jointly track the motion of the target object as well as the distractors or other nearby objects, and can thus be used to more robustly ignore distractor objects.  Optical flow can also track objects over large appearance changes by reasoning about the position of the target object relative to the rest of the scene.  

However, methods for dense optical flow can also make mistakes; incorporating such erroneous information can cause the tracking performance to degrade.  Our main insight is that, to avoid such situations, we should use an estimate of optical flow uncertainty~\cite{FlowNetH} to reason about our confidence in the optical flow estimate.  We develop a novel probabilistic framework that estimates a tracking score for each bounding box based on the optical flow estimates and their uncertainties.  These flow scores are combined with object appearance scores to estimate the new position of the tracked object.
 
 We demonstrate that our method significantly improves tracking performance, when evaluated on the VOT 2016, 2018, and 2019 datasets, compared to the performance of the base tracker that our method builds upon.  Our method is general in that our flow uncertainty tracking scores can be incorporated into any base tracker.  We show that our method improves tracking robustness under distractors and large object appearance changes.  Our contributions include
 \begin{itemize}
     \item A novel end-to-end differentiable method for combining segmentation and flow uncertainty for tracking
     \item Experimental demonstrations that our method outperforms current state of the art trackers
     \item Ablations showing the importance of each component of our method, especially flow uncertainty, for optimal tracking performance
 \end{itemize}

\section{Related Work}
\subsection{2D Instance Tracking}
Since the work of Bolme \etal~\cite{correlation_filter}, correlation filter has been a popular approach for instance tracking. The method trains a filter online and tracks the target by correlating the filter over a search window. Significant efforts has been devoted to improve the performance, such as by learning a multi-channel filter~\cite{multi-channel, KCF}, integrating multi-resolution deep feature maps~\cite{Cont_DCF, HierarchicalConv} and mitigating boundary effects ~\cite{LimitedBoundary, SpatiallyRegularizedCF}.
Recently, instead of learning discriminative filter online, offline learning methods, especially siamese networks~\cite{SiamRPN, DaSiam, SiamMask, Fully_Convolutional_Siamese, GOTURN, DetectToTrack}, have considerably improved performance on 2D instance tracking by using a one-shot detection framework. 

In order to track objects temporally, most trackers incorporate a position penalty to prevent large changes in position from one frame to the next.  This can be achieved using a cosine window penalty~\cite{SiamMask, SiamRPN, DaSiam, Fully_Convolutional_Siamese} or a Gaussian penalty~\cite{RemoveCosine}.  Another approach to incorporate position information more implicitly is to input a search region cropped around the location of the tracked object in previous frame ~\cite{GOTURN, SiamMask, SiamRPN, DaSiam, Fully_Convolutional_Siamese} or to restrict feature correlation to a local neighborhood~\cite{DetectToTrack}. Many previous works take the Bayesian approach for instance tracking, using a Kalman filter~\cite{KalmanFilter, AdaptiveKalmanFilter, 2DHumanBodyTrack} or particle filter~\cite{Condensation, VisualTrackPF, AdaptivePF} to smooth the tracker output over time.  To make the method more robust to distractors, DaSiamRPN~\cite{DaSiam} proposes a distractor-aware module to perform incremental learning during inference time. We show that our approach to avoiding distractors significantly outperforms these approaches. 

Our tracker makes use of a segmentation mask of the tracked object from the previous frame.  To obtain this mask, we use SiamMask~\cite{SiamMask}, which achieves the state-of-the-art tracking performance.  We combine this mask with  uncertainty-aware optical flow to improve tracking performance in the face of distractors and large appearance changes.

\subsection{Optical Flow}
Optical flow has been widely used for video analysis and processing. Traditional methods for optical flow estimation includes variational approaches~\cite{variational_flow}, possibly combined with combinatorical matching~\cite{EpicFlow}. Recently, deep learning based methods~\cite{flownet,flownet2} have obtained state-of-the-art performance for optical flow estimation.  Optical flow has been used to guide feature warping to improve performance of class-level object detection in videos ~\cite{FGFA}. Other work~\cite{seg_by_flow} uses optical flow to identify temporal connections throughout videos, and jointly updates object segmentation with flow models.  In contrast to these applications, we use optical flow to improve tracking performance by estimating how the target object, as well as the other objects in the scene, move over time.

\subsubsection{Tracking with Optical Flow}

Recently, some trackers  \cite{END_TO_END_FLOW_TRACK, DeepMotionFeature, SINT} use optical flow estimation to improve performance on instance tracking. FlowTrack~\cite{END_TO_END_FLOW_TRACK} uses flow to warp features from previous frames to improve the feature representation and tracking accuracy. The warped feature maps are weighted by a spatial-temporal attention module; these feature maps are then input into subsequent correlation filter layers along with feature maps of the current frame. Other work~\cite{DeepMotionFeature} uses optical flow to obtain deep motion features, and then fuses appearance
information with deep motion features for visual tracking. For hand-crafted features, deep image features, and deep motion features, the method separately learns a filter by minimizing the SRDCF~\cite{SRDCF} objective and then averages the filter responses to get final confidence scores. SINT+~\cite{SINT} uses flow to remove motion inconsistent candidates. Specifically, it uses the estimated optical flow to map the locations of the pixels covered by the predicted box in the previous frame to the current frame, and remove the candidate boxes which contain less than 25\% of those pixels. 

However, none of these methods use a segmentation mask or flow uncertainty for tracking; our experiments demonstrate that both of these components are crucial for optimal tracking performance.  We develop a probabilistic framework to use flow uncertainty for tracking.

\subsubsection{Tracking with Uncertainty}
There have been several recent works on estimating confidence in optical flow~\cite{BootstrapFlow, AdaptiveFlow, FlowNetH}. FlowNetH~\cite{FlowNetH} is shown to be able to generate effective uncertainty estimates without the need of sampling or ensembles. As far as we know, these methods have not been used to improve performance of instance tracking.  We propose a new  framework which combines flow uncertainty estimates with appearance scores from a one-shot-detection method; we show that our method can significantly improve tracking robustness and obtains state-of-the-art tracking results.

\section{Background}

\subsection{Siamese Networks for Tracking}
\label{sec:Siamese Networks for Tracking}
Our method is based on the SiamMask~\cite{SiamMask} framework, which is the state-of-the-art method on the VOT tracking benchmarks~\cite{VOT2016,VOT2017,VOT2018,VOT2019}. It consists of siamese subnetwork for feature extraction and a region proposal subnetwork for bounding box proposal generation. The framework scores proposals based on an appearance matching score $d$, a size change penalty $p_s$, and a position change penalty $p_c$. The size change penalty $p_s$ penalizes changes to the size of the bounding box of size $w$ by $h$ from one frame to the next; this is defined as
\begin{align}
    p_s &=  e^{(1-\max(\frac{r}{r'}, \frac{r'}{r}) \cdot \max(\frac{s}{s'}, \frac{s'}{s}))\cdot k_p}
\end{align}
where $s$ and $s'$ are the padded areas $p$ of the proposal box and the bounding box in the previous frame, respectively, given by
\begin{align}
    s^2 &= (w+p) \times (h+p)
\end{align}
and $r$ and $r'$ are the aspect ratios of the proposal box and the bounding box in the previous frame, respectively.
The score $f$ for each proposal is calculated as
\begin{align}
f &= (1-k_c) \cdot  p_s \cdot d + k_c \cdot p_c
\label{eq:total_score}
\end{align}
where $k_c$ and $k_p$ are hyperparameters.  The position penalty $p_c$ is obtained by penalizing the position of the center of the bounding box according to one period of a cosine function, centered at the position of the previous bounding box; the period of this cosine penalty is determined based on the size of the previous bounding box.  The appearance matching score $d$ is given by the output of the one-shot detection network, which matches a template image of the target object to the scene.

SiamMask~\cite{SiamMask} first chooses a bounding box based on the proposal with the highest score $f$.  For the highest scoring proposal box, it then predicts a mask $F_{t}$, thresholds it into a binary mask $\hat{F}_{t}$ using threshold $t_{\text{seg}}$, and outputs the minimum bounding rectangle (MBR) of the binary mask as the final prediction of the location of the tracked object.

\section{Method}
\begin{figure*}[t]
\begin{center}
\includegraphics[width=\linewidth]{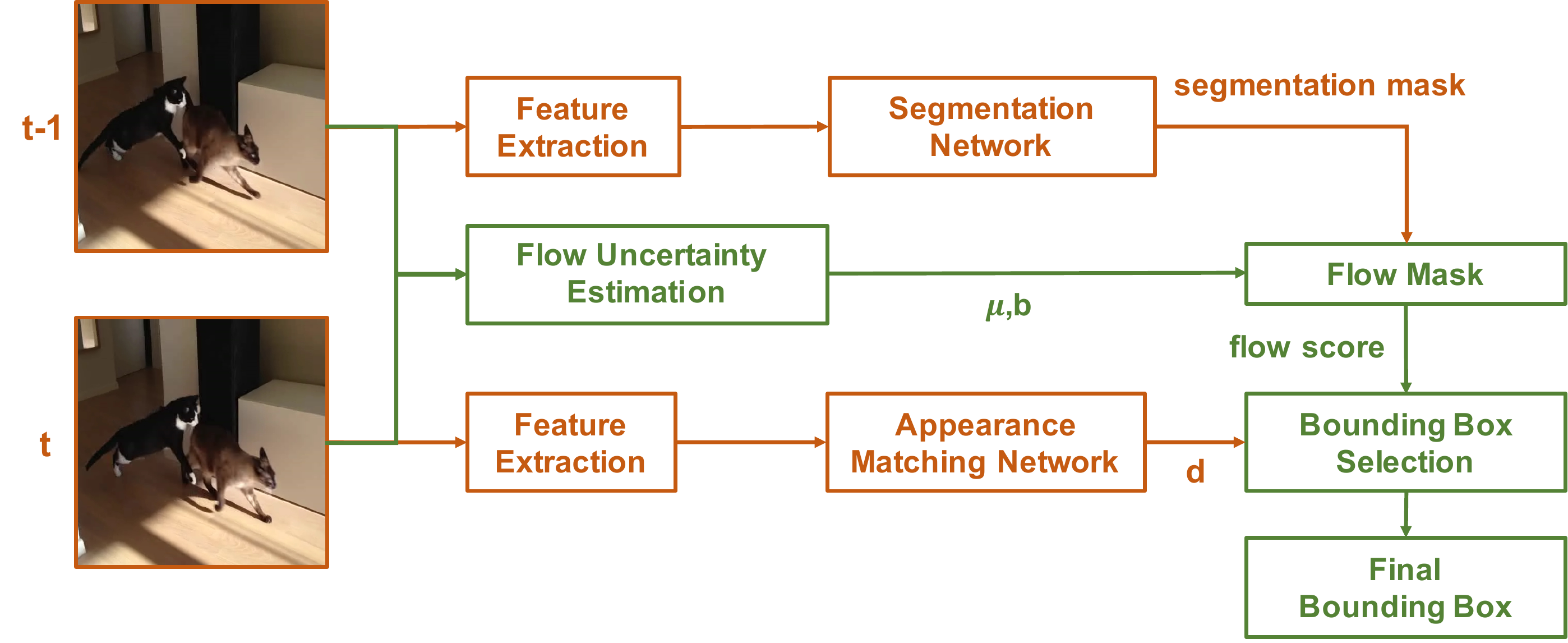}
\end{center}
\caption{Overall pipeline of our method: On top of SiamMask~\cite{SiamMask}(showed in \textcolor{orange}{orange}), we add a module(showed in \textcolor{green}{green}) to compute flow score for each proposal using flow uncertainty estimations and segmentation output from previous frame; our method then combines flow scores with appearance scores to choose a bounding box proposal.}
\label{fig:flowchart}
\end{figure*}

We introduce a new method which improves tracking robustness under distractors and large appearance changes. We visualize our pipeline in Figure~\ref{fig:flowchart}. Our method uses optical flow to estimate the probability of being in part of the foreground for every pixels in a frame, which we call it a ``FlowMask." Based on this probability mask, we assign a flow score for each proposal, which we combine with an appearance score to obtain the final tracking output. The rest of this section explains how our method works in detail.

\subsection{Flow Mask}
\label{sec:Flow Mask}
Our method makes use of previous work for uncertainty-aware dense optical flow estimation~\cite{FlowNetH}  that computes the probability that each pixel $i$ in frame $t$ corresponds to a given pixel location $j$ in frame $t-1$:
$p(c(I_{i,t}) = I_{j,t-1})$
where $c$ maps pixel $I_{i,t}$ to a pixel in frame $t-1$. Given images $I_t$ and $I_{t-1}$, this method predicts the probability of each pixel being part of the foreground as a Laplace distributions, parametrized by flow mean $\mu$ and scale $b$:
\begin{align}
    p(c(I_{i,t}) = I_{j,t-1}) &= \mathcal{L}(I_{j,t-1}-I_{i, t}|\mu, b)
    \label{eq:prod_laplacian}
\end{align}
where the Laplace distributions are defined in the standard manner as
\begin{align}
    \mathcal{L}(u|\mu, b) = \frac{1}{2b}\exp\Big(\frac{-|u - \mu|}{b}\Big)
    \label{eq:laplacian}
\end{align}
For notational convenience, we are omitting the conditioning for these probabilities on the images $I_t$ and $I_{t-1}$.  As we will show, flow uncertainty is crucial for robust tracking. 
Using Eqn.~\ref{eq:prod_laplacian}, we can compute the probability that pixel $I_{i,t}$ corresponds to pixel $I_{j,t-1}$.  We compute the probability that pixel $I_{j,t-1}$ belongs to the target object.  To do so, we use a segmentation-based tracking method~\cite{SiamMask} to obtain a ``segmentation mask", which gives the probability that each pixel $I_{j,t-1}$ belongs to the foreground of the previous bounding box (i.e. the tracked object): $p(I_{j,t-1} \in F_{t-1})$
where $F_{t-1}$ is the set of foreground pixels, i.e. the set of pixels in frame $t-1$ that belong to the tracked object.  

We combine these flow probabilities with the Segmentation Mask probabilities  to estimate the probability that a pixel $I_{i,t}$ in frame $t$ belongs to the tracked object:
\begin{align}
p(I_{i,t} \in F_{t}) = \sum_j p(c(I_{i,t}) = I_{j,t-1}) \, p(I_{j,t-1} \in F_{t-1})
\label{eq:0}
\end{align}
We compute the foreground probability for every pixel $I_{i,t}$ in frame $t$; we refer to the resulting set of probabilities as the ``FlowMask" at frame $t$.  This computation is implemented in a differentiable manner, and could be used in end-to-end trainable pipelines. This idea is further illustrated in Figure~\ref{fig:fig1}.
\\
\noindent\begin{minipage}[b]{0.5\textwidth}
\centering
\includegraphics[width=0.5\textwidth]{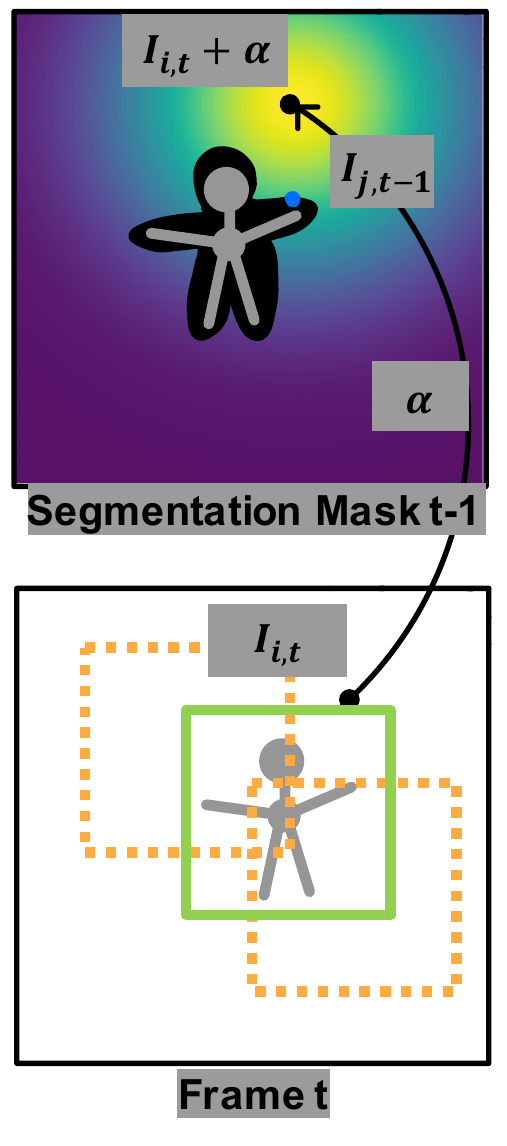}
\captionsetup{width=\textwidth}
\captionof{figure}{Illustration of how flow mask is computed. The \textcolor{green}{green} box is the ground-truth box; the \textcolor{orange}{orange} boxes are proposals; $\alpha_u$ and $\alpha_v$ are the predicted flow mean from frame $t$ to frame $t-1$ in $u,v$ direction; colormap in frame $t-1$ visualizes a Laplace distribution parametrized by predicted flow mean and variance. The blue dot represents a point $I_{j,t-1}$ that belongs to the foreground in frame $t-1$.}
\label{fig:fig1}            
\end{minipage}%
\begin{minipage}[b]{0.5\textwidth}
\centering
\includegraphics[width=\textwidth]{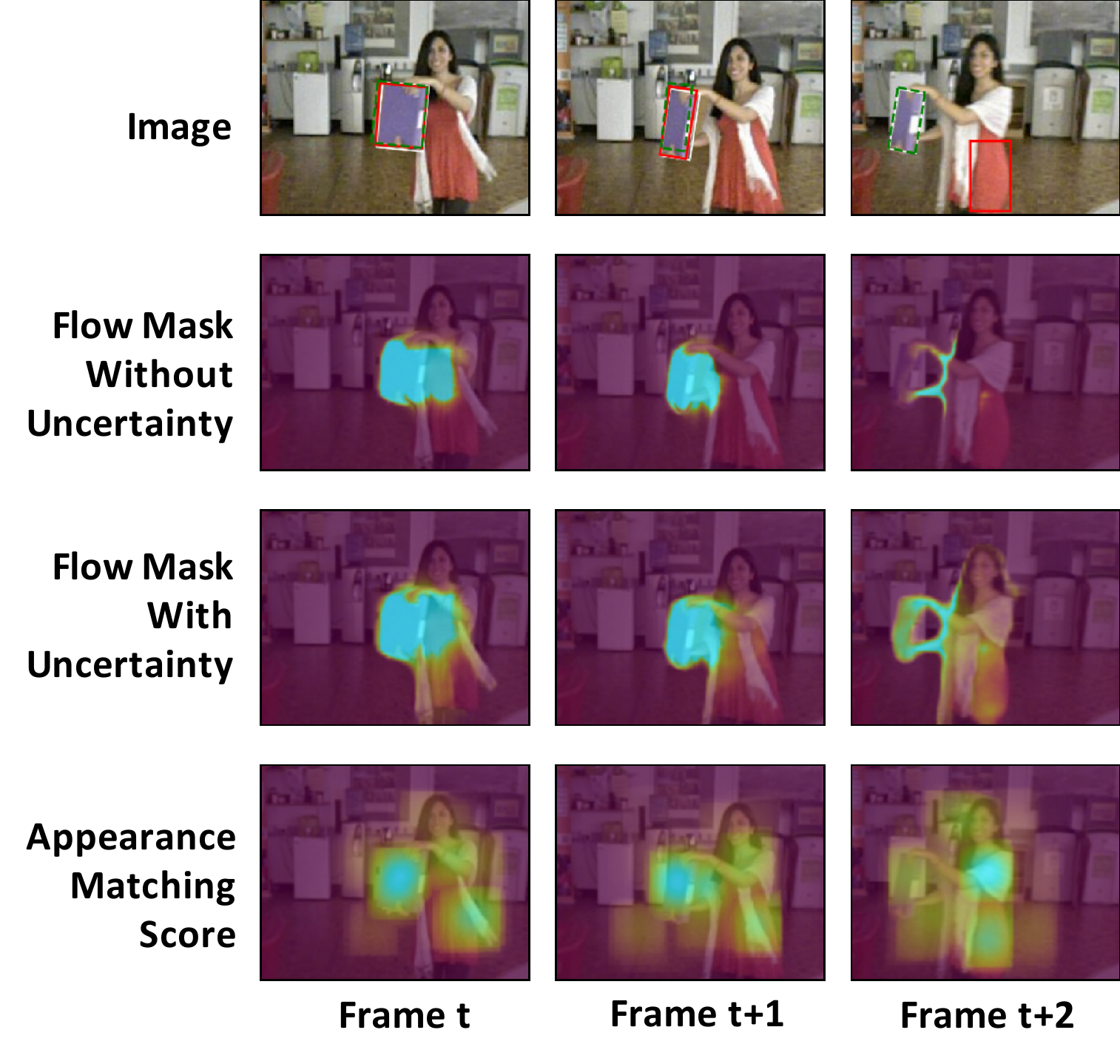}
\captionsetup{width=0.9\textwidth}
\captionof{figure}{Illustration of importance of uncertainty: One example that flow mask with uncertainty successfully tracks the target object but flow mask without uncertainty fails. In this figure, white boxes represent ground-truth boxes, \textcolor{red}{red} boxes represent prediction by using flow mask without uncertainty, \textcolor{green}{green} boxes shows the prediction using flow mask with uncertainty.}
\label{fig:flow_nouncertainty}            
\end{minipage}
\subsection{Flow Score}
After we compute the flow mask at frame $t$, we can compute a flow score for each proposal $box_{(i,t)}$, denote as $f_s(box_{(i,t)})$ by averaging the foreground probabilities for each pixel in the box:
\[f_s(box_{(i,t)}) = \frac{1}{N_{box(i,t)}}\sum_{I_{i,t}\in box_{(i,t)} }p(I_{i,t} \in F_{t})\] where $N_{box(i,t)}$ represents the total number of pixels in $box_{(i,t)}$. However, one discrepancy in this score is that, even though $box_{(i,t)}$ is a rectangle, the object being tracked may not be shaped as a rectangle, which could cause $f_s(box_{(i,t)})$ to be much less than 1.  This variability in $f_s(box_{(i,t)})$ will make it difficult to combine the flow score with the appearance score, as described in Section~\ref{sec:Combining Flow Score with Appearance Score} below.  To deal with this issue, we first compute the number of pixels that are in the thresholded segmentation mask $\hat{F}_{t-1}$ as $N_{\hat{F}_{t-1}}$. Then we compute a $t_{\text{flow},t}$ by dividing $N_{\hat{F}_{t-1}}$ by the number of pixels in previous frame's axis-aligned detection box $box^*_{t-1}$. 
\begin{align}
t_{\text{flow},t} =\frac{N_{\hat{F}_{t-1}}}{N_{box^*_{t-1}}}
\end{align}

This results in a flow score defined as
\begin{align}
    f_s'(box_{(i,t)}) =  \min\Big(\frac{f_s(box_{(i,t)})}{t_{\text{flow},t}},1\Big)
    \label{eq:new_flow_score}
\end{align}

\subsection{Bounding Box Selection}
\label{sec:Combining Flow Score with Appearance Score}
Lastly we combine our flow score with the appearance score from the one-shot detection framework to obtain a motion score for a given proposal box:

\begin{align}
(1-k_f) \cdot p_c 
+ k_f \times f_s'
\label{eq:our_motion_score}
\end{align}
where $k_f$ is a hyperparameter, $p_c$ was described in Section~\ref{sec:Siamese Networks for Tracking}, and $f_s'$ is obtained from Eqn.~\ref{eq:new_flow_score}.  We combine our position penalty $p_c$ from Eqn.~\ref{eq:our_motion_score} with the size penalty $p_s$ and appearance matching score $d$ in Eqn.~\ref{eq:total_score} to obtain the total score for each proposal box. Our entire pipeline is end-to-end differentiable, so we could backprop through our network and learn the value for the different hyperparameters. However, since there are only three hyperparameters,    we proceed as in SiamMask~\cite{SiamMask} to do a hyperparameter searches and find the proposal box with the highest score to obtain the tracking output and estimate a segmentation mask for this box.

\section{Experiments}

\subsection{Implementation Details}
    Our method uses the pretrained SiamMask~\cite{SiamMask} network to obtain the appearance matching score $d$ and to compute the segmentation mask. Since SiamMask~\cite{SiamMask} reports its tracking performace on visual object tracking datasets (VOT 2016, 2018 and 2019), we also report our performance on these three datasets. In SiamMask~\cite{SiamMask}, they perform hyperparameter searches on $k_c \in [0.40,0.43]$ and $k_p \in [0,1]$, and we similarly search in these ranges; we also perform a random hyperparameter search for $k_f \in [0,1]$ for Eqn.~\ref{eq:our_motion_score}.
\subsection{Quantitative Results}

\begin{table*}[h!]
\fontsize{8.5}{12}\selectfont
\centering
\begin{tabular}{||c| c| c| c| c | c | c | c | c | c ||}
 \hline
    &\multicolumn{3}{|c|}{VOT2016}&\multicolumn{3}{|c|}{VOT2018}&\multicolumn{3}{|c||}{VOT2019}\\
     \hline
    & EAO $\uparrow$  & R $\downarrow$ & A $\uparrow$ & EAO $\uparrow$ & R $\downarrow$ & A $\uparrow$ & EAO $\uparrow$ & R $\downarrow$ & A $\uparrow$  \\ [0.5ex] 
 \hline
  ATP~\cite{Kristan2019a} & - & - & -& - &- & - & \textcolor{red}{0.394} & \textcolor{red}{0.291} & \textcolor{red}{0.650} \\
  Our Method & \textcolor{red}{0.47} & \textcolor{red}{0.196} & \textcolor{red}{0.647}& \textcolor{red}{0.41} & \textcolor{red}{0.234} & \textcolor{green}{0.605} & \textcolor{green}{0.306} & \textcolor{green}{0.426} & \textcolor{green}{0.599} \\
 SiamMask~\cite{SiamMask} & \textcolor{green}{0.433} & \textcolor{green}{0.214} & \textcolor{green}{0.639} & \textcolor{green}{0.38} & \textcolor{green}{0.276} & \textcolor{red}{0.609} & 0.283 & 0.467 & 0.596  \\ 
 UInet~\cite{Kristan2019a} & - & - & -& - &- & - & 0.254 & 0.468 & 0.561\\
 SiamMsST~\cite{Kristan2019a} & - & - & -& - &- & - & 0.252 & 0.552 & 0.575\\
 MemDTC~\cite{Memdtc} & 0.297 & 1.310 & 0.5297& 0.2651 &1.5287 & 0.4909& 0.252 & 0.552 & 0.575\\
 CSRDCF~\cite{csrdcf} & 0.338 & 0.85& 0.51 & - & - & - & 0.201 & 0.632 & 0.496\\
 \hline
\end{tabular}
\caption{Results on VOT 2016, VOT2018, and VOT2019. R represents robustness and A represents accuracy. The top three performing trackers are
colored with \textcolor{red}{red} and \textcolor{green}{green} respectively.}
\label{table:results}
\end{table*}
\subsubsection{Evaluation for VOT}

This section includes results on VOT2016~\cite{VOT2016}, VOT2018~\cite{VOT2018}, and VOT2019~\cite{VOT2019}. VOT2016 consists of 60 video sequences. The VOT2017~\cite{VOT2017} challenges replaces the 10 least challenging sequences with new ones. VOT2018 contains the same 60 video sequences as in VOT2017. VOT2019 replaces 20\% of the videos in VOT2018 with new ones.
The performance is evaluated in terms of accuracy (average overlap while tracking successfully), robustness (failure times), and Expected Average Overlap (EAO), which takes account of both accuracy and robustness, as is common for the VOT challenges. 

The results are shown in Table~\ref{table:results}. We compare our method with all the other state-of-the-art trackers that predict both a bounding box for tracking and a segmentation mask for each frame. As can be seen, our method significantly improves over most state-of-the-art baselines in all categories across VOT2016, 2017, and 2018.  In particular, our method builds upon~\cite{SiamMask}, so the improvement should be judged relative to this method.  However, our method for incorporating flow uncertainty into tracking is modular and can be combined with other state-of-the-art tracking methods as well. 

In terms of speed, our methods operates at 5.62 frames per second, or 178ms per frame, including 18ms for SiamMask~\cite{SiamMask} and 60ms for the optical flow computation~\cite{FlowNetH}.


\subsubsection{Ablations}
Our method builds upon SiamMask~\cite{SiamMask} and incorporates flow uncertainty based on the previous frame's predicted segmentation mask to improve performance. To further investigate the importance of different components of our method, as well as the effectiveness of different approximations that we make, we conduct the following ablations: 
\paragraph*{Importance of Optical Flow}
We first investigate the importance of using optical flow for tracking, rather than the approach taken by several recent papers of using a cosine~\cite{SiamMask, SiamRPN, DaSiam, Fully_Convolutional_Siamese} or Gaussian penalty~\cite{RemoveCosine} to penalize large motions from the previous frame.  To analyze this, we note that our method for optical flow uses a Laplacian distribution, as shown in Equations~\ref{eq:prod_laplacian} and~\ref{eq:laplacian}.  Thus, to evaluate the importance of optical flow, we replace the estimated flow distribution with a constant Laplacian, with 0 mean $\mu= (0,0)$ and fixed scale parameters $b$.  This ablation is referred to as ``Ours minus Flow" (Ours - Flow) in Table~\ref{table:ablation_analysis}.  As can be seen, using a constant Laplacian distribution (rather than optical flow) leads to no improvement over the baseline SiamMask~\cite{SiamMask}.
\paragraph*{Importance of Optical Flow with Uncertainty}
For the next ablation, we probe the importance of utilizing uncertainty estimates for optical flow in tracking.  To evaluate this, we fix the Laplacian scale parameters $b$ to a constant. The result is shown in Table~\ref{table:ablation_analysis} as ``Ours - Uncertainty."  Although the result is better than our baseline, it still has a large performance gap with our method. As a qualitative analysis, Figure~\ref{fig:flow_nouncertainty} shows one example where our method successfully tracks the target object but Ours-Uncertainty fails due to errors in the flow estimate under large object motion and perspective changes. This shows how the uncertainty increases the tracking robustness.



\paragraph*{Importance of Segmentation Mask}
In this ablation, we investigate the importance of using a segmentation mask for the computation of the flow mask. In our method, we use SiamMask~\cite{SiamMask} to predict a segmentation mask for the previous frame.  We then combine the probability of being in the foreground with the flow probability, as shown in Equation~\ref{eq:0}. We probe the importance of having a segmentation mask by replacing it with a bounding box.  We analyze using both an axis-aligned bounding box (ALB) and a minimum bounding rectangle (MBR) mask (see SiamMask~\cite{SiamMask} for details). The result is shown in Table~\ref{table:ablation_analysis} as ``Ours - SegMask (ALB)" and  ``Ours - SegMask (MBR)". As we can see, using the minimum bounding rectangle instead of a segmentation mask results in no improvement over the baseline. On the other hand, using the mask obtained using axis-align box improve upon baseline but still is not as effective as our method. 

\subsubsection{SiamMask+Flow Rejection}
Last, we compare to an additional baseline that also uses optical flow to improve tracking.
Following SINT+~\cite{SINT}, we evaluate using optical flow to filter out motion inconsistent candidates, and try to use this ``flow rejection'' method to improve  the  SiamMask~\cite{SiamMask}. Specifically, we use flow to warp the pixels covered by the predicted box in the previous frame.  We  then remove all proposals in the current frame that contain less than 25\% of the warped pixels (this is similar to the procedure from SINT+~\cite{SINT}). We refer to this experiment as ``SiamMask plus flow rejection'' (SiamMask + FlowRej) in Table~\ref{table:ablation_analysis}. As we can see, using flow rejection does not improve the performance compared to the baseline SiamMask~\cite{SiamMask}. This degradation in performance, especially in robustness, is likely due to occasional errors in the flow estimation.  This supports our claim about the important of flow uncertainty estimation for robust tracking.

\begin{table}[h!]
\fontsize{8.5}{12}\selectfont
\centering
\begin{tabular}{||c| c| c| c||} 
 \hline
    &\multicolumn{3}{|c||}{VOT2018}\\
     \hline
    & EAO$\uparrow$  & Robustness$\downarrow$ & Accuracy$\uparrow$ \\ [0.5ex] 
 \hline
 Ours & \textbf{0.41} & \textbf{0.234} & 0.605\\
 Ours - Flow &  0.38 & 0.276 & 0.609\\
 Ours - Uncertainty & 0.383 & 0.262 & 0.610\\
 Ours - SegMask (ALB) & 0.388 & 0.267 & \textbf{0.614}\\
 Ours - SegMask (MBR) & 0.372 & 0.253 & 0.593\\
 SiamMask~\cite{SiamMask} + FlowRej & 0.361 & 0.290 & 0.613\\
  SiamMask~\cite{SiamMask} & 0.38 & 0.276 & 0.609 \\ 
\hline
\end{tabular}
\caption{Ablation Analysis. Ours-Flow uses identity flow; Ours-Uncertainty uses fixed variance; Ours-SegMask (ALB) replaces segmentation mask with an axis-aligned bounding box; Ours-SegMask (MBR) replaces segmentation mask with a minimum bounding rectangle.}
\label{table:ablation_analysis}
\end{table}

\subsection{Qualitative Analysis}
Our method effectively improves tracking robustness under distractors and large object appearance changes. To better illustrate the effect of our method, we analyze our results  qualitatively.
In Figure~\ref{fig:all}, we visualize three cases where the state-of-the-art tracker SiamMask~\cite{SiamMask} fails but our method is able to successfully keep track of the target objects. For each case, we visualize the position penalty that SiamMask~\cite{SiamMask} uses, the appearance matching score (i.e appearance\_score) produced by the one-shot-detection network, and the flow mask introduced in this work in Section~\ref{sec:Flow Mask}. Figure~\ref{fig:all} shows two categories of challenging tracking scenarios:

\paragraph*{Distractors} 
One type of common failure case occurs when there are distractor objects in the background that are similar in appearance or category to the object being tracked.  An example is shown in Figure~\ref{fig:all}(b), in which the target object runs across another person in the background. In this case, the appearance matching score (from the one-shot-detection network of SiamMask) is high for both people.  The position penalty is also not useful in this case due to the fast motion of the target object.  Thus, if we only rely on the appearance matching score and the position penalty, we would track the distractor instead of the target object, as illustrated by the red detection box (output by SiamMask). 

Nevertheless, the flow mask successfully tracks the target object. Since the flow mask is a probabilistic estimate based on the predicted segmentation mask of the previous frame, it is able to focus precisely on the target object; additionally, because we incorporate flow uncertainty, our method is also robust to small errors in the estimated flow. 


\paragraph*{Large Appearance Changes}
Another challenging problem in tracking is large appearance changes. In Figure~\ref{fig:all}(a), the image of the target object becomes blurry under large camera motion. In this scenario, the appearance matching network predicts similar confidence for many areas in the image. The position penalty also fails because the position of the large change in the object position on the image due to the fast camera motion.
Similarly, in Figure~\ref{fig:all}(c), the position penalty is also not effective due to the fast motion of the target object. In this case, the deformation of the target object (a bird) also causes the appearance matching score to be uncertain, leading to a failure from SiamMask.

However, in both cases, our proposed flow mask is still able to track the target object. These examples illustrate that our method is robust to large appearance changes, such as blurry images and deformations, as well as fast moving objects.


\subsection{Detailed Analysis on VOT2018}
\begin{figure*}
\begin{center}
\includegraphics[width=0.8\linewidth]{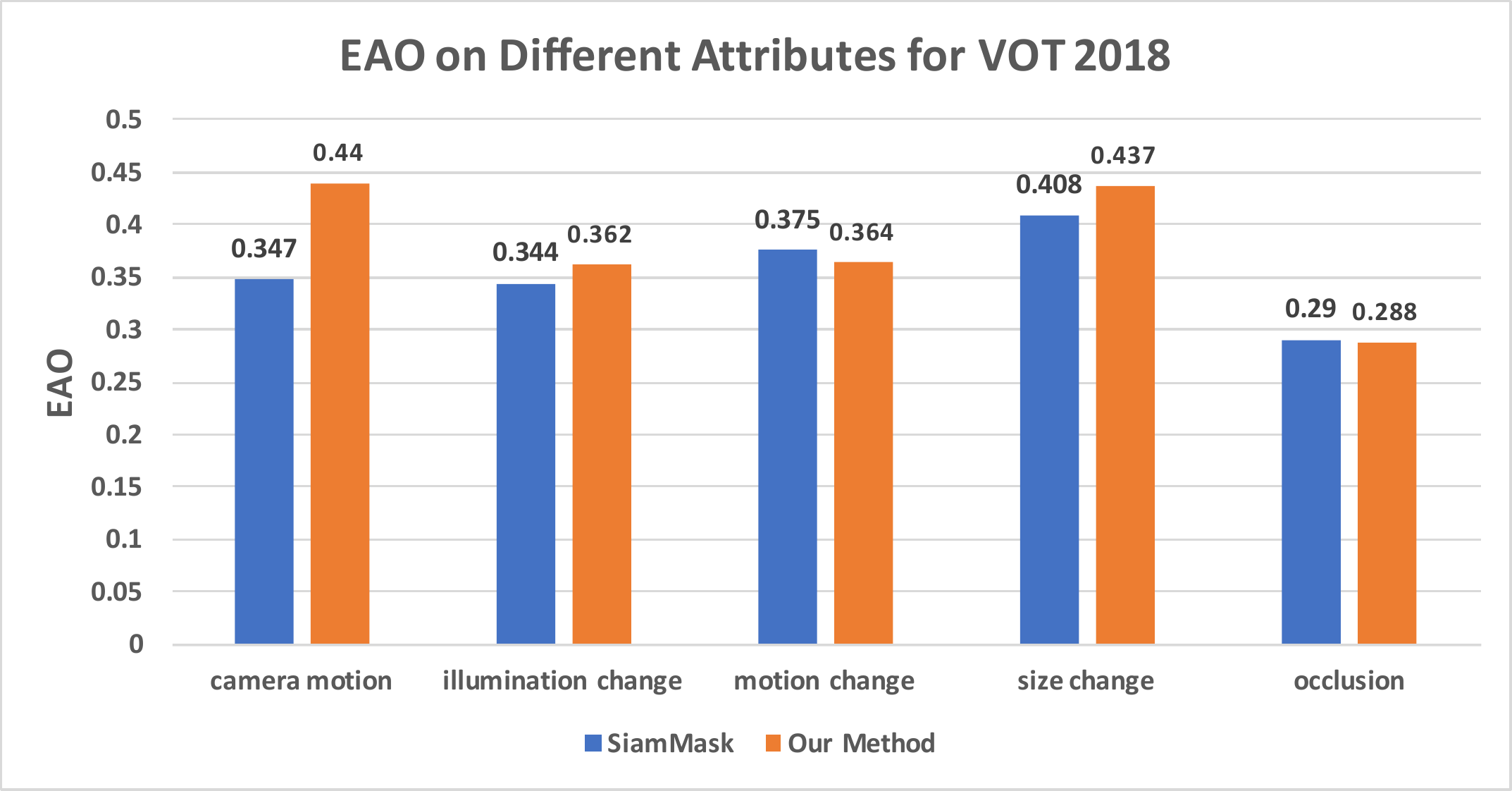}
\end{center}
   \caption{Results breakdown on VOT 2018 for different visual attributes. We compare the EAO under these five attributes of our method with the baseline SiamMask~\cite{SiamMask}. }
\label{fig:attributes}
\end{figure*}

To better understand the effect of using our method, we perform in-depth analysis on the VOT 2018 dataset. In the VOT 2018 dataset, each frame is manually labeled with five visual attributes that reflect a particular challenge: (i) camera motion, (ii) motion change, (iii)  size change, (iv) illumination change, (v) occlusion. In case that a frame doesn't correspond to any of those five attributes, it is labeled as "non-degraded". Those labels enable us to analyze the benefits of our method while focusing only on the frames that contain a given attribute.  

The results are shown in Figure~\ref{fig:attributes}, in which we compare our method to the SiamMask~\cite{SiamMask} baseline that we build on top of. As can be seen, our method significantly improves the tracker's performance under camera motion, and we also see modest improvements under size change and illumination change. Our method performance slightly worse than SiamMask~\cite{SiamMask} under motion change and occlusion. 

\begin{figure*}
\begin{center}
\includegraphics[width=\linewidth]{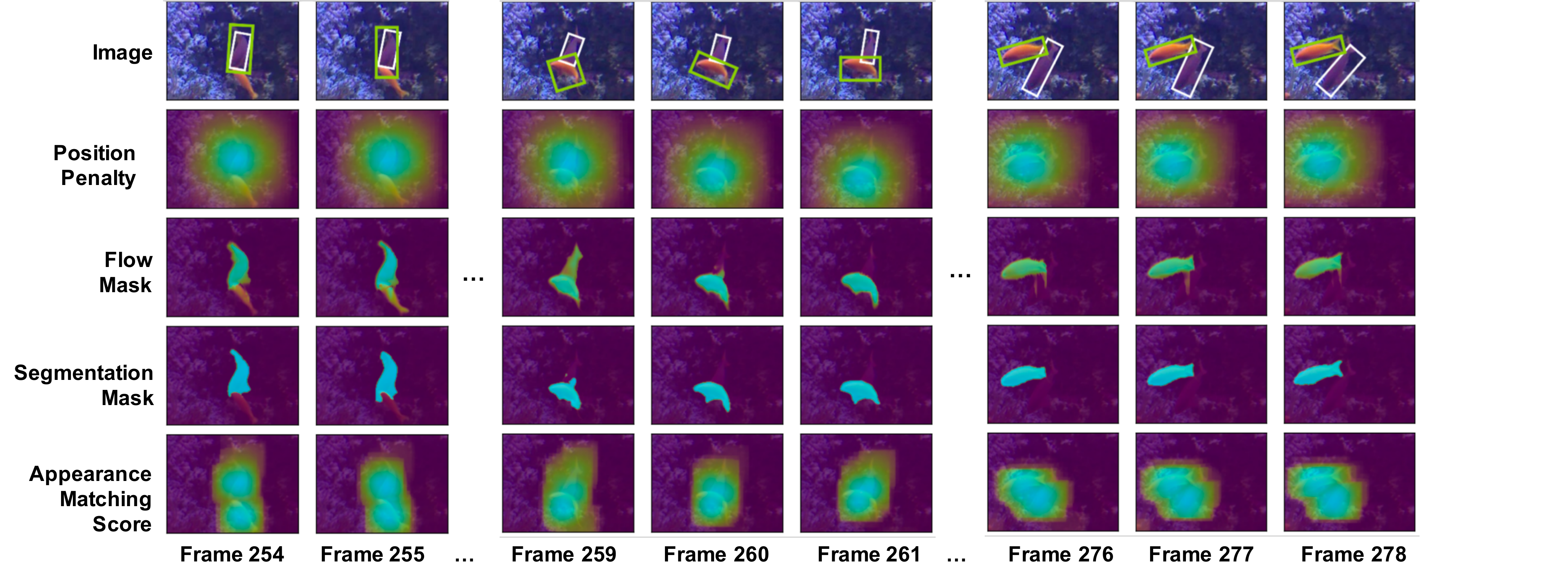}
\end{center}
   \caption{Illustration of a failure case due to occlusion: when there is an occlusion, the predicted segmentation mask and flow mask tends to drift to the distractor. In this figure, white boxes represent the ground-truth , \textcolor{green}{green} boxes show the prediction from our method.}
\label{fig:fish2_failure}
\end{figure*}

In Figure~\ref{fig:fish2_failure}, we visualize one example of a failure case due to occlusion. In this case, the target object gets occluded by another similar looking distractor. We find that, when an occlusion occurs, the predicted segmentation mask tends to also mask the distractor; thus it will mislead the calculation of FlowMask in the following frames. Eventually both FlowMask and the segmentation mask would have high confidence for the distractor. Thus tracker would drift and track the distractor instead. In SiamMask~\cite{SiamMask} baseline, it fails similarly in this case. However, since the location of two objects don't change too much, eventually baseline method recovers by the help of position penalty. In our case, the use of flow mask prevent us from recovering from a failure in this case.
\section{Conclusions}
In this paper, we introduce a novel probabilistic framework that combines appearance and flow uncertainty for tracking. We show that our method, when evaluated on Visual Object Tracking datasets, significantly improves the performance of a state-of-the-art tracker. Ablation experiments show the importance of each component of our framework, such as the use of flow uncertainty and warping a segmentation mask. We hope that our work can be insightful to future research on robust tracking under distractor objects and large object appearance changes.
\newpage
\bibliographystyle{splncs}
\bibliography{egbib}

\begin{thebibliography}{10}

\bibitem{SiamRPN}
Li, B., Yan, J., Wu, W., Zhu, Z., Hu, X.:
\newblock High performance visual tracking with siamese region proposal
  network.
\newblock In: The IEEE Conference on Computer Vision and Pattern Recognition
  (CVPR). (2018)

\bibitem{DaSiam}
{Zha}, Y., {Wu}, M., {Qiu}, Z., {Dong}, S., {Yang}, F., {Zhang}, P.:
\newblock Distractor-aware visual tracking by online siamese network.
\newblock IEEE Access \textbf{7} (2019)  89777--89788

\bibitem{SiamMask}
{Wang}, Q., {Zhang}, L., {Bertinetto}, L., {Hu}, W., {Torr}, P.H.S.:
\newblock Fast online object tracking and segmentation: A unifying approach.
\newblock In: 2019 IEEE/CVF Conference on Computer Vision and Pattern
  Recognition (CVPR). (2019)  1328--1338

\bibitem{Fully_Convolutional_Siamese}
Bertinetto, L., Valmadre, J., Henriques, J.F., Vedaldi, A., Torr, P.H.S.:
\newblock Fully-convolutional siamese networks for object tracking.
\newblock In: ECCV 2016 Workshops. (2016)  850--865

\bibitem{FlowNetH}
Ilg, E., Cicek, O., Galesso, S., Klein, A., Makansi, O., Hutter, F., Brox, T.:
\newblock Uncertainty estimates and multi-hypotheses networks for optical flow.
\newblock In: The European Conference on Computer Vision (ECCV). (2018)

\bibitem{correlation_filter}
Bolme, D., Beveridge, J., Draper, B., Lui, Y.:
\newblock Visual object tracking using adaptive correlation filters.
\newblock (2010)  2544--2550

\bibitem{multi-channel}
{Galoogahi}, H.K., {Sim}, T., {Lucey}, S.:
\newblock Multi-channel correlation filters.
\newblock In: 2013 IEEE International Conference on Computer Vision. (2013)
  3072--3079

\bibitem{KCF}
Henriques, J.F., Caseiro, R., Martins, P., Batista, J.:
\newblock High-speed tracking with kernelized correlation filters.
\newblock IEEE Transactions on Pattern Analysis and Machine Intelligence
  \textbf{37} (2015)  583--596

\bibitem{Cont_DCF}
Danelljan, M., Robinson, A., Shahbaz~Khan, F., Felsberg, M.:
\newblock Beyond correlation filters: Learning continuous convolution operators
  for visual tracking.
\newblock In: ECCV. (2016)

\bibitem{HierarchicalConv}
Ma, C., Huang, J.B., Yang, X., Yang, M.H.:
\newblock Hierarchical convolutional features for visual tracking.
\newblock 2015 IEEE International Conference on Computer Vision (ICCV) (2015)
  3074--3082

\bibitem{LimitedBoundary}
Galoogahi, H.K., Sim, T., Lucey, S.:
\newblock Correlation filters with limited boundaries.
\newblock 2015 IEEE Conference on Computer Vision and Pattern Recognition
  (CVPR) (2014)  4630--4638

\bibitem{SpatiallyRegularizedCF}
Danelljan, M., H{\"a}ger, G., Khan, F.S., Felsberg, M.:
\newblock Learning spatially regularized correlation filters for visual
  tracking.
\newblock 2015 IEEE International Conference on Computer Vision (ICCV) (2015)
  4310--4318

\bibitem{GOTURN}
Held, D., Thrun, S., Savarese, S.:
\newblock Learning to track at 100 fps with deep regression networks.
\newblock In: ECCV. (2016)

\bibitem{DetectToTrack}
Feichtenhofer, C., Pinz, A., Zisserman, A.:
\newblock Detect to track and track to detect.
\newblock 2017 IEEE International Conference on Computer Vision (ICCV) (2017)
  3057--3065

\bibitem{RemoveCosine}
Li, F., Wu, X., Zuo, W., Zhang, D., Zhang, L.:
\newblock Remove cosine window from correlation filter-based visual trackers:
  When and how.
\newblock ArXiv \textbf{abs/1905.06648} (2019)

\bibitem{KalmanFilter}
Başar, T.:
\newblock A new approach to linear filtering and prediction problems.
\newblock (2001)

\bibitem{AdaptiveKalmanFilter}
Weng, S.K., Kuo, C.M., Tu, S.K.:
\newblock Video object tracking using adaptive kalman filter.
\newblock J. Visual Communication and Image Representation \textbf{17} (2006)
  1190--1208

\bibitem{2DHumanBodyTrack}
Jang, D.S., Jang, S.W., Choi, H.I.:
\newblock 2d human body tracking with structural kalman filter.
\newblock Pattern Recognition \textbf{35} (2002)  2041--2049

\bibitem{Condensation}
Isard, M., Blake, A.:
\newblock Condensation—conditional density propagation for visual tracking.
\newblock International Journal of Computer Vision \textbf{29} (1998)  5--28

\bibitem{VisualTrackPF}
Hossain, K., Lee, C.W.:
\newblock Visual object tracking using particle filter.
\newblock 2012 9th International Conference on Ubiquitous Robots and Ambient
  Intelligence (URAI) (2012)  98--102

\bibitem{AdaptivePF}
Li, X., Lan, S., Jiang, Y., Xu, P.:
\newblock Visual tracking based on adaptive background modeling and improved
  particle filter.
\newblock 2016 2nd IEEE International Conference on Computer and Communications
  (ICCC) (2016)  469--473

\bibitem{variational_flow}
Horn, B., Schunck, B.:
\newblock Determining optical flow.
\newblock In: Artificial Intelligence. (1981)

\bibitem{EpicFlow}
{Revaud}, J., {Weinzaepfel}, P., {Harchaoui}, Z., {Schmid}, C.:
\newblock Epicflow: Edge-preserving interpolation of correspondences for
  optical flow.
\newblock In: 2015 IEEE Conference on Computer Vision and Pattern Recognition
  (CVPR). (2015)  1164--1172

\bibitem{flownet}
{Dosovitskiy}, A., {Fischer}, P., {Ilg}, E., {Häusser}, P., {Hazirbas}, C.,
  {Golkov}, V., v.~d. {Smagt}, P., {Cremers}, D., {Brox}, T.:
\newblock Flownet: Learning optical flow with convolutional networks.
\newblock In: 2015 IEEE International Conference on Computer Vision (ICCV).
  (2015)  2758--2766

\bibitem{flownet2}
{Ilg}, E., {Mayer}, N., {Saikia}, T., {Keuper}, M., {Dosovitskiy}, A., {Brox},
  T.:
\newblock Flownet 2.0: Evolution of optical flow estimation with deep networks.
\newblock In: 2017 IEEE Conference on Computer Vision and Pattern Recognition
  (CVPR). (2017)  1647--1655

\bibitem{FGFA}
{Zhu}, X., {Wang}, Y., {Dai}, J., {Yuan}, L., {Wei}, Y.:
\newblock Flow-guided feature aggregation for video object detection.
\newblock In: 2017 IEEE International Conference on Computer Vision (ICCV).
  (2017)  408--417

\bibitem{seg_by_flow}
Tsai, Y.H., Yang, M.H., Black, M.J.:
\newblock Video segmentation via object flow.
\newblock In: The IEEE Conference on Computer Vision and Pattern Recognition
  (CVPR). (2016)

\bibitem{END_TO_END_FLOW_TRACK}
Zhu, Z., Wu, W., Zou, W., Yan, J.:
\newblock End-to-end flow correlation tracking with spatial-temporal attention.
\newblock In: The IEEE Conference on Computer Vision and Pattern Recognition
  (CVPR). (2018)

\bibitem{DeepMotionFeature}
Gladh, S., Danelljan, M., Khan, F.S., Felsberg, M.:
\newblock Deep motion features for visual tracking.
\newblock 2016 23rd International Conference on Pattern Recognition (ICPR)
  (2016)  1243--1248

\bibitem{SINT}
Tao, R., Gavves, E., Smeulders, A.W.M.:
\newblock Siamese instance search for tracking.
\newblock 2016 IEEE Conference on Computer Vision and Pattern Recognition
  (CVPR) (2016)  1420--1429

\bibitem{SRDCF}
Danelljan, M., H{\"a}ger, G., Khan, F.S., Felsberg, M.:
\newblock Learning spatially regularized correlation filters for visual
  tracking.
\newblock 2015 IEEE International Conference on Computer Vision (ICCV) (2015)
  4310--4318

\bibitem{BootstrapFlow}
Kybic, J., Nieuwenhuis, C.:
\newblock Bootstrap optical flow confidence and uncertainty measure.
\newblock Computer Vision and Image Understanding \textbf{115} (2011)
  1449--1462

\bibitem{AdaptiveFlow}
Kondermann, C., Kondermann, D., J{\"a}hne, B., Garbe, C.S.:
\newblock An adaptive confidence measure for optical flows based on linear
  subspace projections.
\newblock In: DAGM-Symposium. (2007)

\bibitem{VOT2016}
Kristan, M., Leonardis, A., Matas, J., Felsberg, M., Pflugfelder, R.,
  \v{C}ehovin Zajc, L., Vojir, T., H\"{a}ger, G., Luke\v{z}i\v{c}, A.,
  Fernandez, G.:
\newblock The visual object tracking vot2016 challenge results.
\newblock Springer (2016)

\bibitem{VOT2017}
Kristan, M., Leonardis, A., Matas, J., Felsberg, M., Pflugfelder, R.,
  \v{C}ehovin Zajc, L., Vojir, T., H\"{a}ger, G., Luke\v{z}i\v{c}, A.,
  Eldesokey, A., Fernandez, G.:
\newblock The visual object tracking vot2017 challenge results (2017)

\bibitem{VOT2018}
Kristan, M., Leonardis, A., Matas, J., Felsberg, M., Pfugfelder, R.,
  \v{C}ehovin Zajc, L., Vojir, T., Bhat, G., Lukezic, A., Eldesokey, A.,
  Fernandez, G., et~al.:
\newblock The sixth visual object tracking vot2018 challenge results (2018)

\bibitem{VOT2019}
Kristan, M., Matas, J., Leonardis, A., Felsberg, M., Pflugfelder, R.,
  Kamarainen, J.K., \v{C}ehovin Zajc, L., Drbohlav, O., Lukezic, A., Berg, A.,
  Eldesokey, A., Kapyla, J., Fernandez, G.:
\newblock The seventh visual object tracking vot2019 challenge results (2019)

\bibitem{Kristan2019a}
Kristan, M., Matas, J., Leonardis, A., Felsberg, M., Pflugfelder, R.,
  Kamarainen, J.K., \v{C}ehovin Zajc, L., Drbohlav, O., Lukezic, A., Berg, A.,
  Eldesokey, A., Kapyla, J., Fernandez, G.:
\newblock The seventh visual object tracking vot2019 challenge results (2019)

\bibitem{Memdtc}
{Yang}, T., {Chan}, A.B.:
\newblock Visual tracking via dynamic memory networks.
\newblock IEEE Transactions on Pattern Analysis and Machine Intelligence (2019)
   1--1

\bibitem{csrdcf}
Lukezic, A., Voj{\'i}r, T., Zajc, L.C., Matas, J.E.S., Kristan, M.:
\newblock Discriminative correlation filter tracker with channel and spatial
  reliability.
\newblock International Journal of Computer Vision \textbf{126} (2017)
  671--688

\end{thebibliography}

\end{document}